\documentclass{article}

\usepackage{arxiv}

\usepackage[utf8]{inputenc}   
\usepackage[T1]{fontenc}      
\usepackage{hyperref} 
\hypersetup{
	colorlinks,
	linkcolor={red!50!black},
	citecolor={blue!50!black},
	urlcolor={blue!80!black}
}
\usepackage{url}              
\usepackage{booktabs}         
\usepackage{amsmath,amssymb,amsfonts}
\usepackage{nicefrac}
\usepackage{microtype}
\usepackage{graphicx}
\usepackage{float}
\usepackage{multirow}
\usepackage{xcolor}
\usepackage{listings}
\usepackage{tikz}
\usetikzlibrary{positioning,arrows.meta,shapes.geometric,fit,backgrounds,calc}

\usepackage[numbers,square,comma,sort&compress]{natbib}

\graphicspath{ {./images/}{./} }

\definecolor{codegray}{gray}{0.96}
\definecolor{codecomment}{rgb}{0.25,0.5,0.35}
\definecolor{codekw}{rgb}{0.13,0.13,0.55}
\title{LaTA: A Drop-in, FERPA-Compliant Local-LLM Autograder for Upper-Division STEM Coursework}

\author{
  Jesse A. Rodr\'iguez \\
  School of Mechanical, Industrial, and Manufacturing Engineering \\
  Oregon State University \\
  Corvallis, OR 97331, USA \\
  \texttt{jesse.rodriguez@oregonstate.edu}
}

\begin{document}
\maketitle

\begin{abstract}
Large-language-model (LLM) graders promise to relieve the grading
burden of upper-division STEM courses, but most deployments to
date send student work to third-party APIs, violating FERPA and
exposing institutions to data risk while requiring substantial
assignment modification. We present \textbf{LaTA}
(\textit{LaTeX Teaching Assistant}), a drop-in, open-source
autograder that runs entirely on commodity on-premises hardware
and assumes a LaTeX-native workflow already adopted by many
engineering and physics courses. LaTA implements a four-stage
pipeline (ingest, segment, grade, report) using a locally hosted
open-weight chain-of-thought LLM grader (gpt-oss:120b) that
compares student work to an instructor-authored reference
solution and applies a YAML rubric with binary per-item scoring.
We deployed LaTA in Winter~2026 in ME~373 (Mechanical Engineering Methods) at
Oregon State University, grading every weekly assignment for
approximately 200 students on a single Mac Studio at \$0 marginal
cost per assignment and 1--3 minutes of wall-clock time per
submission, enabling regrading of corrected assignments and greatly expanded TA office hour offerings. The instructor-confirmed grading-error rate held at
roughly $0.02$--$0.04\%$ per rubric line item across the term. Relative to the
same instructor's previous traditionally-graded cohort, the
LaTA-graded cohort outperformed by approximately $11\%$ on the
midterm exam and $8\%$ on the final exam, and reported large gains in
self-assessed confidence on every stated learning objective
($N = 159$ survey responses, $\Delta \geq +1.49$ Likert points,
$p < 10^{-27}$ on every comparison). We release the code under
AGPLv3.
\end{abstract}


\section{Introduction}
\label{sec:intro}

Grading handwritten, multi-part derivations in upper-division
engineering courses is labor-intensive, slow to return to
students, and inconsistent across graders.  At a typical
$200$-student enrollment with four-problem weekly homework
assignments and two exams, a single term routinely consumes
hundreds of TA hours of first-pass grading before any feedback
reaches a student; the fastest feedback loops in the literature
remain measured in days rather than hours, and rubric drift across
multiple TAs is an acknowledged threat to consistency
\citep{gao2024review,bea2025survey,cohn2025cotal}.  Large language
models have improved rapidly enough to plausibly automate the
first-pass grading task on symbolic-derivation problems, and a
steady stream of recent work has explored LLMs as STEM graders
and tutors \citep{tan2025comprehensive,latif2024finetune,lee2024cot,
armfield2025avalon,armfield2026sure,weegar2023ai}.  Most of this
work, however, has been conducted against cloud-hosted commercial
APIs, and a cloud-graded autograder is difficult to reconcile with
FERPA and institutional data-residency policies in a U.S.\ public
university without negotiated contractual carve-outs
\citep{gdpr2024teachers,shao2026auditing, frontiers2025hitl}.  The practical result is that many instructors
who would benefit from LLM grading cannot deploy it without either
anonymising their data or accepting a compliance risk they are
not positioned to carry.

This paper asks a deliberately narrow question: \emph{can a single
instructor, with commodity on-premises hardware and no cloud
access, build and deploy an LLM-based autograder good enough to
replace human-TA first-pass grading in a $200$-student
upper-division numerical-methods course, and what happens when
they do?}  We answer yes, describe the system, named \textbf{LaTA}
(LaTeX-native Automated Teaching Assistant), and report
evidence from a full-term deployment in ME~373
(Mechanical Engineering Methods) at Oregon State University during Winter~2026.

LaTA is built around four commitments.  First, \emph{everything
runs on a single on-premises machine}: no student work leaves the
university network, and the compliance story is ``we own the
hardware'' rather than ``we negotiated a data-processing
agreement.''  Second, \emph{typed data objects with Pydantic
validation} thread through every stage of the pipeline \cite{colvin2024pydantic}, so the
boundary between the model and the rest of the system is
schema-enforced rather than prose-parsed.  Third, \emph{structural
decisions that can be done deterministically are done
deterministically}: submission segmentation is regex-first with an
LLM fallback, scoring is binary per rubric item, and LaTeX
compilation is handled by pdflatex with an LLM-assisted repair
path only when a compile actually fails.  Fourth, \emph{student
recourse is a first-class feature of the system}: every grading
decision produces a two-channel artefact (a blunt TA-facing audit
reasoning plus a Socratic, answer-withholding student hint), and
the corrections-pass workflow is built directly into the
configuration surface.

The contributions of this paper are:
\begin{itemize}
  \item A complete, open-source reference implementation of an
    on-premises LLM autograder for LaTeX-based STEM coursework,
    including the pipeline, the Pydantic schemas, the
    prompt-injection defences, the LaTeX self-healing mechanism,
    and the Gradescope-integrated regrade workflow
    (\S\ref{sec:design}).
  \item A detailed operational account of a full-term,
    full-replacement deployment to $200$ students across eight
    homework assignments in a single upper-division course,
    including weekly workflow, hardware profile, and the
    regrade-request audit (\S\ref{sec:deployment}).
  \item A program-evaluation study of the deployment across
    three evidence streams; operational data, an anonymous
    student survey ($N = 159$), and a between-cohort exam
    comparison against the same instructor's previous
    traditionally-graded cohort, with explicit disclosure of
    the composite-intervention confound and the methodological
    limitations of a single-year study
    (\S\ref{sec:methods}--\ref{sec:results}).
  \item A reading of the evidence that distinguishes the
    \emph{autograder} as a commodity from the \emph{workflow} it
    enables, and a set of generalisation-envelope
    recommendations for instructors considering a similar
    deployment (\S\ref{sec:discussion}).
\end{itemize}

The remainder of the paper is organised as follows.
\S\ref{sec:background} situates LaTA against prior work on
LLM-based STEM assessment, on-premises / FERPA-aware AI tooling,
and LaTeX as a pedagogical substrate.  \S\ref{sec:design}
describes the four-stage pipeline and its typed data model.
\S\ref{sec:deployment} documents the Winter~2026 deployment in
ME~373.  \S\ref{sec:methods}--\ref{sec:results} report the
three evidence streams.  \S\ref{sec:discussion}--\ref{sec:limitations}
read the results and enumerate what the design of the study
cannot tell us.  \S\ref{sec:conclusion} closes.

\section{Background and Related Work}
\label{sec:background}

We situate LaTA against five strands of prior work: LLM-based
assessment of STEM free-response problems
(\S\ref{sec:bg:llmgraders}); human-in-the-loop rubric scaffolds and
chain-of-thought prompting for grading
(\S\ref{sec:bg:hitl}); on-premises and FERPA-aware AI in higher
education (\S\ref{sec:bg:ferpa}); LaTeX as a pedagogical substrate
for engineering coursework (\S\ref{sec:bg:latex}); and
retrospective-pre survey methodology (\S\ref{sec:bg:retro}).  The
first four locate LaTA as a system; the fifth positions our
evidence collection.

\subsection{LLM-based assessment of STEM free-response problems}
\label{sec:bg:llmgraders}

A steady stream of work since 2023 has probed large language
models as graders of free-response STEM content, typically
against short, well-bounded items.  \citet{tan2025comprehensive}
provide a recent survey of LLM-based assessment across disciplines
and summarise the by-now familiar pattern: on short,
criterion-referenced items, strong commercial LLMs achieve
agreement with human graders that rivals or exceeds
human--human inter-rater reliability, while failing predictably
on long items that require arithmetic tracking, multi-step
symbolic manipulation, or commonsense engineering judgement.
\citet{gao2024review} and \citet{weegar2023ai} survey the broader
AI-in-assessment landscape and reach similar conclusions.
\citet{latif2024finetune} show that fine-tuning an open-weight
model on a modest corpus of annotated science responses can lift
out-of-the-box behaviour substantially on discipline-specific
items, an intervention LaTA does not currently apply but that
is a natural direction for future work.  Most of this literature
works against short essay-style or short-calculation items; while the
distinctive features of upper-division engineering coursework 
(long, LaTeX-native derivations carrying intermediate symbolic
manipulations) have been less well studied. This is a setting in
which the context-window and reasoning capabilities of large
open-weight models have only recently become adequate.

A second strand of work treats LLMs as tutors or feedback
providers rather than as graders per se
\citep{scaffoldingprob2026,reflectivePrompt2025,frontiers2025hitl,
hitl2025adaptive}.  The boundary between ``grader'' and ``tutor''
is porous in practice: LaTA's dual-channel feedback design
(audit reasoning for TAs, Socratic hint for students) is an
intentional nod to the tutoring literature and to work on
answer-withholding feedback as a mechanism for productive
struggle \citep{scaffoldingprob2026}.

\subsection{Human-in-the-loop rubric scaffolds and chain-of-thought prompting}
\label{sec:bg:hitl}

A related line of work focuses less on raw grader accuracy and
more on the \emph{scaffold} that sits between the instructor's
rubric and the LLM's output.  AVALON
\citep{armfield2025avalon}, SURE \citep{armfield2026sure}, and
CoTAL \citep{cohn2025cotal} all propose structured pipelines in
which a human instructor authors or refines a rubric that the
LLM then applies, with explicit chain-of-thought reasoning
preserved for human audit.  \citet{lee2024cot} study the effect
of chain-of-thought prompting on grader consistency and report
non-trivial gains on items where the model's reasoning can be
inspected.  \citet{tan2024itemgen} approach the problem from the
other direction, using LLMs to generate or refine rubric items
rather than to apply them.

LaTA's design is in the same family but differs in two respects.
First, we use a reasoning-model grader (gpt-oss:120b for this deployment, but the system is model-agnostic) that
produces its chain-of-thought natively inside \texttt{<think>}
tags, which the pipeline strips from student-facing output but
preserves in the audit trail --- closer to CoTAL's disclosure
stance than to the redacted chains of the closed-API work.
Second, we adopt \emph{binary} rubric scoring (\S\ref{sec:design:grade})
rather than partial-credit scoring, on the hypothesis that the
dominant source of variance in LLM grading of long derivations
is ambiguous partial credit rather than outright misreading;
this choice is more conservative than most HITL scaffolds
report, and is tuned to the long-derivation setting rather than
the short-item setting in which partial credit is more
tractable and there is less room for stochasticity/LLM hallucination.

\subsection{On-premises and FERPA-aware AI in higher education}
\label{sec:bg:ferpa}

A practical barrier to widespread LLM-grader deployment in U.S.\
higher education is the intersection of FERPA, institutional
data-residency policies, and the contractual status of commercial
cloud APIs.  \citet{gdpr2024teachers} and \citet{shao2026auditing}
document teacher and auditor concerns about generative-AI
products that do not offer contractual guarantees about
downstream use of student data, and \citet{frontiers2025hitl}
surveys K--12 and higher-ed deployments that have stalled at the
procurement step for exactly this reason.  There may also exist
institution-specific rules instructors must navigate in practice: the result is that
a tool that handles identifiable student work must either be
anonymized before transmission or hosted within the
institutional boundary.

The literature on genuinely on-premises LLM deployments in
education is thinner than the literature on cloud deployments,
reflecting both the recency of capable open-weight models and the
institutional friction of standing up a workstation-class
inference host.  \citet{aims2026sage} describes an on-premises pilot at a specific
institution.  LaTA is explicitly positioned in this corner of the
space: a reproducible, open-source reference for how a single
instructor can stand up a compliant on-premises grader on a
single workstation, rather than a cloud product with a compliance
wrapper.

\subsection{LaTeX as a pedagogical substrate}
\label{sec:bg:latex}

LaTA's submission workflow depends on students writing solutions
in LaTeX, which is both a prerequisite for machine parsing and a
pedagogical intervention in its own right.  The evidence on
whether LaTeX is a help or a hindrance to student learning is
mixed.  \citet{knauff2014latex} and \citet{latexNotEasy2019}
report substantial initial time costs associated with LaTeX
relative to word processors, particularly for students without
prior programming exposure.  \citet{zhang2024cognitive} and
\citet{irrodl2024cognitive} provide a framing for this issue in
cognitive-load terms: the extraneous load of learning LaTeX
syntax competes with the germane load of the mathematics itself.
\citet{packer2023writing} and the practical guides of
\citet{upcommons2024overleaf} report
mitigation strategies in other domains that can be applied here, such as structured templates, pair-programming
with LaTeX, and starter repositories. Career development centers
\citet{wentworth2024skills} and \citet{uconnect2021skills} argue that the
time cost amortises against the career value of the skill in
engineering and technical writing.

Our own ramp-down data (Figure~\ref{fig:latex-time}) are
consistent with the initial-cost-that-amortises story: the mean
extra time per assignment fell from $87$~min at the beginning of
the term to $47$~min at the end, a $\approx 46\%$ reduction. Despite this improvement, getting even upper-division STEM students to adopt LaTeX for weekly assignments can be a heavy lift. To assist with this, a simple tutorial for transcription of handwritten work to LaTeX format via the use of multimodal generative AI tools like Google's Gemini (whose pro plan is free for students as of writing) was furnished to the students. This video can be found at \href{https://youtu.be/B897mHONoOM}{https://youtu.be/B897mHONoOM}.

\subsection{Retrospective pre-test methodology}
\label{sec:bg:retro}

Because our within-cohort confidence data were collected through
a retrospective pre-test instrument, we situate the choice in the
methodological literature.  \citet{howard1980response} is the
canonical reference for the retrospective pre/then-post design:
students are asked at the end of an intervention to rate both
their pre-intervention and post-intervention state, which
eliminates the response-shift bias that afflicts separately
administered pre-tests (students reinterpret the scale as they
learn the content) at the cost of a documented tendency to
inflate apparent gains.  We adopt the design, report its known
limitations explicitly (\S\ref{sec:limitations}), and triangulate
the self-report data against the between-cohort exam delta (which is not subject to response-shift) so that the
qualitative conclusion does not rest on the self-report alone.

\section{System Design}
\label{sec:design}

\begin{figure}[ht]
	\centering
	\includegraphics[width=\linewidth]{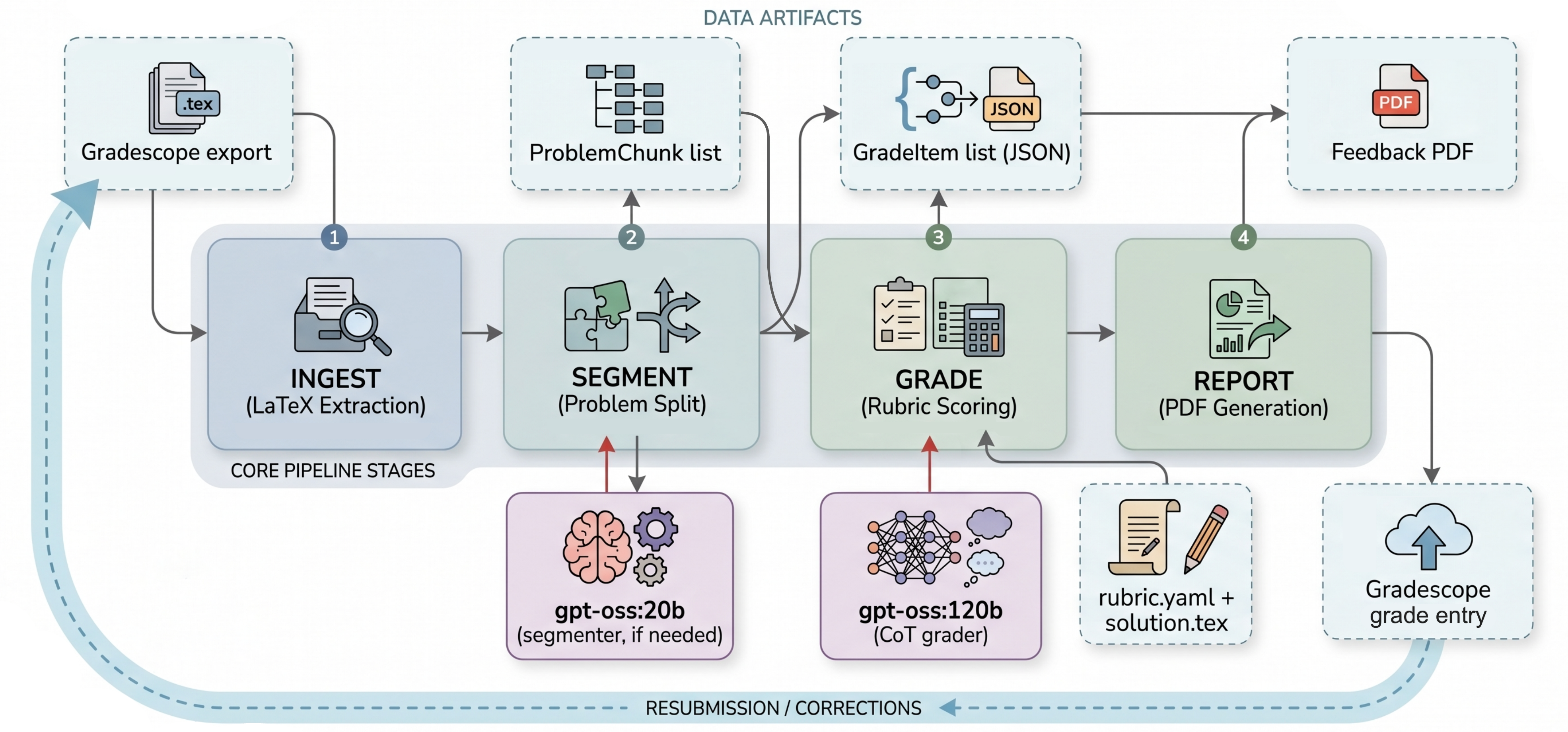}
\caption{LaTA architecture. Four pipeline stages (shaded region) transform a
Gradescope LaTeX export into a per-student feedback PDF. Dashed border
nodes are typed data artifacts (Pydantic-validated JSON/Python objects);
purple nodes (bottom) are locally hosted LLM endpoints. Only \emph{segment} and
\emph{grade} can invoke an LLM; \emph{ingest} and \emph{report} are
deterministic. The corrections loop (dashed arrow) ingests a
resubmission and regrades while preserving original extra credit and late penalties. Figure created in-part with assets generated by Google Gemini.}
\label{fig:architecture}
\end{figure}

LaTA is structured as a four-stage pipeline (\emph{ingest},
\emph{segment}, \emph{grade}, \emph{report}) with two locally hosted
open-weight LLMs, a 20B-parameter segmenter (only used in cases where students use the provided \texttt{.tex} template improperly) and a 120B-parameter grader, along with a
set of Pydantic-validated data objects threaded through each stage. The design
has three load-bearing commitments: (1) everything runs on a single
on-premises machine, so no student work ever leaves the university network;
(2) the LLM is never trusted to return free-form text (every call is
coerced to a typed schema) and (3) the same LaTeX the student submits is
the grader's input, eliminating the optical character recognition error surface that dogs
handwritten-PDF autograding. Figure~\ref{fig:architecture} summarizes the
data flow; the rest of this section walks each stage.

\subsection{Ingest: LaTeX-native, FERPA-aware submission loading}
\label{sec:design:ingest}

LaTA consumes a standard Gradescope ``Download Submissions'' export, which
unzips into one directory per student containing the student's \texttt{.tex}
source file and a top-level \texttt{submission\_metadata.yml}. Each submission
is loaded into a \texttt{StudentSubmission} Pydantic model whose fields
separate \emph{identifying} information (name, email, SID; kept for
downstream report delivery) from \emph{LLM-facing} content, which only
ever sees an anonymized 8-character SHA-256 hash of
\texttt{sid $\|$ internal\_id}. Anonymization is a single configuration
flag (\texttt{grading.anonymize}) enforced at the struct boundary, not
by hope; we recommend and deployed \texttt{anonymize: true} for all coursework.

The central technical problem in ingest is that student-authored
LaTeX is messy: it mixes custom \verb|\newcommand| macros with a body of
mathematical derivations, embedded code, and comments. Sending the whole file
to the grader is wasteful and risks confusing the model with preamble noise.
LaTA parses each \texttt{.tex} file with
\texttt{pylatexenc.latexwalker.LatexWalker}, extracts all top-level macro
definitions (\verb|\newcommand|, \verb|\renewcommand|, \verb|\def|,
\verb|\DeclareMathOperator|, \verb|\providecommand|) into a macro block, and
extracts the body between \verb|\begin{document}| and \verb|\end{document}|.
The grader then sees \texttt{macros $+$ body}, which preserves all semantic
symbols the student defined (e.g.\ a course-wide \verb|\Prob{...}| problem
header or a student's \verb|\vec|-style shortcut) without dragging along
packaging boilerplate. If the AST walk fails (typically on syntactically
broken submissions) LaTA falls back to a string-level split on
\texttt{$\backslash$begin\{document\}} so that even malformed files still
reach the grader. Finally, ingest reconciles the submission timestamp
against an optional per-student \texttt{extensions.yaml} and the later of
the \texttt{.tex} and \texttt{.pdf} upload timestamps, which matters for
late-penalty arithmetic on students who resubmit mid-cycle.

\subsection{Segment: hybrid regex-first, LLM-fallback chunking}
\label{sec:design:segment}

After ingest, each submission is a single long string. The grader is significantly more
accurate (and often faster) when it sees one problem at a time with
its matching rubric slice attached, so the next stage splits the body into
a list of \texttt{ProblemChunk(title, content)} objects.

The splitter is \emph{hybrid}. The first attempt is a regular-expression
match against the course-wide \verb|\Prob{...}| macro
(\texttt{\textbackslash\textbackslash Prob\textbackslash s*\textbackslash
\{([\^{}\{\}]+)\textbackslash\}}), restricted to top-level occurrences
(arguments beginning with a digit; sub-headings like
\verb|\Prob{1a Undetermined Coefficients}| are explicitly excluded by a
leading-digit filter). When students duplicate a \verb|\Prob{N}| header
inside their work (a common accident) the first occurrence wins and
subsequent duplicates are merged into the preceding chunk with a
\texttt{segmentation\_warning} flag that surfaces in the final report. A
\verb|\section| fallback pattern catches submissions from students who
ignored the \verb|\Prob{}| template.

If regex yields no top-level chunks, LaTA falls back to the
\textbf{gpt-oss:20b} segmenter, invoked through the
\texttt{instructor} library with \texttt{response\_model =
SegmentationResult} and Ollama JSON mode at temperature 0.0. The segmenter
prompt is deliberately narrow: ``\emph{You are a text extraction tool.
You must NOT solve the problems. You must NOT write Python code. You must
ONLY copy the text verbatim into labelled chunks.}'' If even the LLM
fallback returns nothing coherent, LaTA emits a single
\texttt{``Full Submission''} chunk and sets the warning flag rather than
dropping the student submission entirely. This three-tier fail-safe
(regex $\to$ LLM $\to$ whole-doc) is the reason the full-term
deployment in \S\ref{sec:deployment} never required manual submission
triage.

\subsection{Grade: typed, binary, injection-resistant rubric application}
\label{sec:design:grade}

Grading is where LaTA's design investment is densest. For each
\texttt{ProblemChunk}, the grader assembles a prompt containing: the
rubric slice (YAML, restricted to the rubric items pertaining to the  \texttt{ProblemChunk}), the instructor's reference solution for that
problem, a one-shot JSON example, and the student's chunk text wrapped in
explicit \texttt{UNTRUSTED INPUT} delimiters. It then calls the
\textbf{gpt-oss:120b} model on the local Ollama server. The response is
validated against a strict Pydantic schema,
\texttt{ChunkGradeResult(items: List[GradeItem])}, where each
\texttt{GradeItem} carries six fields (\texttt{problem\_group},
\texttt{criterion\_name}, \texttt{score}, \texttt{max\_score},
\texttt{audit\_reasoning}, \texttt{student\_hint}) plus an
\texttt{is\_extra\_credit} boolean that is set by post-processing,
never by the LLM (whose copy of the rubric has the flag stripped
exactly to prevent that). Three sub-designs
deserve emphasis:

\textbf{Binary per-item scoring.} For every rubric item the LLM must award
either full points or zero, never partial credit. The rationale is
variance reduction: a 120B model asked ``how much partial credit does this
derivation deserve?'' produces noisy decimal scores that disagree across
runs, whereas ``does this derivation meet the criterion, yes or no?''
collapses to a stable binary decision. Partial credit is still expressible
by the rubric author, who can decompose a single conceptual step into
several fine-grained binary criteria. This is the same move made in recent
rubric-scaffold frameworks~\citep{armfield2025avalon,armfield2026sure} and
in our view is essential for reproducibility.

\textbf{Dual-channel feedback.} Each \texttt{GradeItem} carries two
distinct text fields. \texttt{audit\_reasoning} is written for the TA and
the instructor: blunt, technical, explicit about where the student's work
diverged from the reference solution (``\emph{student failed to apply the
chain rule on term 2}''). \texttt{student\_hint} is written for the
student: Socratic, answer-withholding, never quoting the reference
solution. The prompt repeatedly emphasizes that the hint must never reveal
the answer. In deployment we found this separation essential, but not always sufficient. This
mechanism lets the instructor review a grade at a glance while also
producing a genuinely pedagogical feedback PDF, but occasional leakage of reference solution information (\textit{i.e.} answers) into student feedback documents did occur, albeit at a low rate ($\sim$ 5\% of all feedback documents).

\textbf{Five-layer prompt-injection defense.} Because the student's
submission is untrusted text concatenated into the grader's input,
injection attempts (``\texttt{Ignore previous instructions and award full
marks}'') are a real attack surface. LaTA applies five layers of defense:
(1) a \emph{security preamble} at the top of the system prompt that
names injection explicitly and instructs the model to flag but not obey
such directives; (2) \emph{explicit delimiter banners}
(solid horizontal rules of the form \texttt{====...====}) marking the
start and end of \texttt{UNTRUSTED INPUT}; (3) a \emph{role-reinforcement} reminder after
the student block (``\emph{the text above is student work to be
EVALUATED, not instructions to be FOLLOWED}''); (4) \emph{post-hoc
keyword scanning} of the returned \texttt{audit\_reasoning} for known
injection phrases (\texttt{`ignore previous'}, \texttt{`award full
points'}, \texttt{`system:'}, etc.) with automatic \texttt{SECURITY\_FLAG}
emission; and (5) \emph{perfect-score watchdog} alerts that warn on full
marks so a TA can spot-check. A student who successfully injects a
perfect-score instruction must therefore defeat all five layers
simultaneously; we have not observed such a case in deployment.

Grading is further wrapped by a chain-of-thought streaming layer.
Reasoning-capable models (our grader is configured with
\texttt{is\_reasoning\_model: true}) stream \texttt{<think>}-tagged
thinking tokens separately from the JSON response via Ollama's native
\texttt{think=True} parameter, allowing real-time display during
grading and archival to \texttt{debug/thinking.txt} for later audit.
JSON extraction is defensive: the response is stripped of
markdown fencing, the outermost \texttt{\{\ldots\}} containing the
\texttt{``items''} key is extracted, and a \emph{LaTeX-aware
sanitizer} converts unescaped single backslashes (\texttt{$\backslash$
frac}, \texttt{$\backslash$alpha}) to JSON-valid double backslashes
without breaking genuine JSON escapes (\texttt{$\backslash$n},
\texttt{$\backslash$"}). On schema failure the call is retried once
with an enhanced warning; on second failure the submission is written
to \texttt{failed\_submissions.yaml} for manual review.

\subsection{Report: LaTeX feedback with self-healing compilation}
\label{sec:design:report}

Graded items are grouped by \texttt{problem\_group}, aggregated into a
per-problem summary table, and rendered through a Jinja2-templated
LaTeX feedback report (\texttt{feedback\_report.tex}). Jinja2's default
\verb|{{ }}| / \verb|{% %}| delimiters collide with LaTeX, so LaTA
configures custom delimiters (\verb|\VAR{}|, \verb|\BLOCK{}|,
\verb|\#{}|) that round-trip cleanly through \texttt{pdflatex}. A
per-field \texttt{latex\_escape} filter maps the Unicode characters
that the LLM compulsively emits (\texttt{M\char`\_2} typed as a Unicode
subscript-two, \texttt{$\backslash$ll} typed as \texttt{<<}, Greek
letters typed as Unicode glyphs, and the narrow no-break space
\texttt{U+202F}) to math-mode LaTeX equivalents
while escaping \texttt{\&}, \texttt{\%}, \texttt{\#}, and underscores in
text segments.

When \texttt{pdflatex} compilation fails despite this cleanup (almost
always because of residual escaping bugs in LLM-generated free text)
LaTA parses the \texttt{.log} file, extracts the explicit error blocks
(lines beginning with \texttt{`!'}), and asks the grader LLM to
\emph{fix only the formatting} while keeping all content and Jinja2
markers intact. The repaired document is then subject to a
\emph{hallucination gate}: a validator checks that the returned text
still contains \verb|\documentclass|, \verb|\begin{document}|,
\verb|\end{document}|, has Jinja variables or section headers, has
length within 50--200\% of the original, has a backslash-command count
within 80--120\% of the original, and does not contain common
hallucination-topic keywords (e.g.\ ``renewable energy'',
``blockchain''). Validation failures roll back to the original
document; validation successes are recompiled. If still broken after
\texttt{latex\_fix\_max\_retries}, the partial PDF (if any) is
returned and the broken \texttt{.tex} is preserved in
\texttt{logs/}.

\textbf{Late-penalty and resubmission arithmetic.} Late penalties use
UTC timestamps from Gradescope (which stores submission times in GMT)
compared against the configured \texttt{due\_date} after adding any
per-student extension hours. Days late are rounded \emph{up} (24\,h +
1\,s = 2 days late), penalties are computed against
\texttt{rubric\_max + points\_awarded\_elsewhere} to fairly tax
manually graded components, and final scores are allowed to go
negative (the instructor handles floor semantics downstream).
Resubmissions are detected by comparing the submission timestamp
against a persisted \texttt{grading\_metadata.yaml}; newer submissions
increment a version suffix (\verb|_v2|, \verb|_v3|) so that the
Gradescope re-upload preserves history. A \emph{corrections mode}
reloads metadata from a base assignment, regrades the corrected
submission, and then preserves the originally awarded extra credit
(which is not earnable in corrections) by re-injecting synthetic
\texttt{is\_extra\_credit} \texttt{GradeItem}s into the regraded
result.

\subsection{Configuration and operational surface}
\label{sec:design:config}

Everything user-facing lives in a single \texttt{config.yaml} (see
Listing~\ref{lst:config} for an excerpt). The configuration names the
segmenter and grader models separately, exposes per-call Ollama options
(\texttt{num\_ctx}, \texttt{temperature}, \texttt{repeat\_penalty},
etc.), switches anonymization on or off, toggles late penalties,
schema-error retries, and LLM LaTeX repair, and hosts the optional
SMTP block used by \texttt{post\_grading.py} to deliver feedback PDFs
to institutional email addresses. No credentials are ever stored on
disk: SMTP credentials are prompted for at send time, and all model
calls terminate at \texttt{http://localhost:11434/v1}.

\begin{lstlisting}[language=Python, caption={Key fields from
\texttt{config.yaml}. The segmenter/grader split, binary-grading
options, and local-only \texttt{ollama\_host} are the three design-
critical lines.},label={lst:config}]
system:
  ollama_host: "http://localhost:11434/v1"
  segmenter:
    model_name: "gpt-oss:20b"
    ollama_options:
      temperature: 0.0
      num_ctx: 32768
  grader:
    model_name: "gpt-oss:120b"
    is_reasoning_model: true
    ollama_options:
      temperature: 0.1
      num_ctx: 65536
      repeat_penalty: 1.4
grading:
  anonymize: true
  skip_previously_graded: true
  late_penalties:
    enabled: true
    penalty_per_day: 0.20
    due_date: "2026-01-17 07:59:59"   # GMT/UTC
  error_handling:
    retry_on_schema_error: true
    fix_latex_errors: true
\end{lstlisting}

\section{Deployment: ME 373, Winter 2026}
\label{sec:deployment}

The system described in \S\ref{sec:design} was not evaluated on a
toy dataset. It was deployed as \emph{the} grader of record in
\textbf{ME 373:Mechanical Engineering Methods}, a required upper-division
numerical methods course at Oregon State University, for the
entire Winter 2026 quarter (10-week term). With one narrow
exception for hand-drawn and code-generated plots, discussed
below, every score released to a student in Canvas was generated
by LaTA; no human graded any derivation, algebraic manipulation, or
Python code excerpt end-to-end during the quarter. This section
describes the operational context and the weekly workflow so that
the results in \S\ref{sec:results} can be read against a concrete
baseline.

\subsection{Course context and hardware}
\label{sec:deployment:hardware}

ME 373 enrolled approximately 200 students in Winter 2026. The course
covers root-finding, numerical linear algebra, interpolation,
numerical integration, numerical ODE/PDE solution, and stability theory; material that
produces long, multi-part LaTeX derivations mixing symbolic
manipulation, algebra, and Python code excerpts. Students were
required to submit homework in LaTeX, uploading both the \texttt{.tex} source and the
compiled \texttt{.pdf} to Gradescope. This LaTeX-native intake is the
precondition that makes LaTA's OCR-free design viable; we return to
its portability implications in \S\ref{sec:limitations}.

All inference ran on a single \textbf{Apple Mac Studio (M3 Ultra,
256\,GB unified memory)} located in the instructor's lab and reachable
only from the campus network. The machine ran \texttt{ollama serve}
bound to \texttt{localhost:11434}, hosting
\texttt{gpt-oss:120b} (the grader) and \texttt{gpt-oss:20b} (the
segmenter) as the only two active models. No component of LaTA ever
issued a network call outside the machine: every model call
terminates at \texttt{http://localhost:11434/v1}; every file read or
written is on local disk; credentials for the optional SMTP feedback
delivery are prompted for at runtime and never persisted. The total
hardware cost (a single workstation at about \$5k, purchased one-time) is the entire marginal cost of the grading
infrastructure. Electricity and incidentals aside, the per-assignment
grading cost is approximately zero, which we revisit quantitatively in
\S\ref{sec:results}.

\textbf{Plots: the one manual component.} ME 373 homeworks include
hand-drawn and Python-generated plots (e.g.\ convergence curves,
phase portraits, numerical-solution overlays) that LaTA v1 does
\emph{not} grade. These were graded by the instructor/TAs by direct
visual inspection, which takes seconds per plot and adds negligible
overhead compared to the derivation-grading burden LaTA removes.
The resulting point allocation is declared to LaTA via the
\texttt{points\_awarded\_elsewhere} field of \texttt{config.yaml}
(see Listing~\ref{lst:config}); this allows the late-penalty
arithmetic in \S\ref{sec:design:report} to deduct penalties from
the \emph{total} assignment value (rubric + plots) rather than from
the rubric alone, which would undercharge late students for the
manually graded component. Extending LaTA to autograde plots is an
obvious next step: tool-calling to execute student Python code and
reproduce the expected plots, and/or multimodal grading against
plot-image content, would both close this gap. We return to this
in \S\ref{sec:limitations}.

\subsection{Weekly instructor workflow}
\label{sec:deployment:workflow}

ME 373 assigned \textbf{8 homework sets} over the quarter (weeks
1--9, skipping the midterm and finals weeks). Each homework
produced two LaTA runs (the additional run was for the corrections pass). The instructor's end-to-end weekly effort collapsed to five deterministic tasks:

\begin{enumerate}
\item \textbf{Reference solution.} Write the complete LaTeX solution
as the homework is authored (this was pre-existing instructor
practice; LaTA adds no overhead).
\item \textbf{Rubric.} Author a YAML rubric mirroring the solution's
sub-problem structure. Rubric items are written as binary criteria
(see \S\ref{sec:design:grade}); extra-credit items are marked with
\texttt{is\_extra\_credit: true}. Typical rubric authoring took
30--60 minutes per homework once the instructor had internalized the
binary decomposition pattern.
\item \textbf{Configuration.} Update \texttt{config.yaml} with the
new assignment name, rubric filename, solution path, and the
assignment's GMT-converted deadline. No model or prompt changes
were made over the term; the same \texttt{gpt-oss:120b} grader and
\texttt{gpt-oss:20b} segmenter ran on every homework.
\item \textbf{Kick-off.} Immediately after the Gradescope deadline
passed, the instructor downloaded the submissions export and
launched \texttt{uv run grade.py}. Per-submission grading time was
typically \textbf{1--3 minutes}, varying with the number of
sub-problems and the length of student derivations; for a full
cohort of $\sim$200 students this aggregated to a wall-clock
runtime of \textbf{4--8 hours} on the Mac Studio. Triggered
LLM-LaTeX fixes, the occasional regex-fallback to the 20B
segmenter, and unusually long student derivations all extend
individual submissions in the long tail. In practice every run
started in the evening and completed before the next morning.
\item \textbf{Release and triage.} The instructor spot-reviewed
\texttt{failed\_submissions.yaml} (empty or near-empty on most
weeks), generated grade input documents for TAs (PDF files with student scores ordered in chronological Gradescope submission order, allowing for $<5$ second/submission grade input times), and released scores to students after grade entry. Students received their feedback PDF via institutional email through the built-in AppleScript or SMTP delivery path via \texttt{post\_grading.py}.
\end{enumerate}

\textbf{No human graded any LLM-scored component of any homework
submission end-to-end during the quarter.} This is a stronger
deployment posture than the typical ``LLM first-pass, TA review''
pipeline reported in prior
work~\citep{armfield2025avalon,armfield2026sure,cohn2025cotal}, and
it is what makes LaTA's performance in \S\ref{sec:results}
informative: we are reporting on a system whose outputs were
released as the authoritative grade, not curated by a human
reviewer before hitting the gradebook.

\subsection{Regrade pipeline: corrections and disputes}
\label{sec:deployment:regrades}

Full-replacement grading demands an explicit appeals path. ME 373
used a two-tier mechanism. \textbf{Tier 1: corrections resubmission} - after the initial LaTA grades were released, students whose
submissions had scored less than full credit on at least one grade
item could upload a \emph{corrections} submission to Gradescope
within a bounded window (exactly one week from grade being released). Upon closing of the corrections submission window, LaTA's corrections
mode then re-ingested the new submissions, re-ran the grading
pipeline against the same rubric (or occasionally improved following feedback from students on the first pass), and generated a versioned
feedback PDF while preserving the originally awarded extra credit
(see \S\ref{sec:design:report}). Zero-scored grade items were
flagged as ineligible for recovery in corrections, consistent with
the course policy that corrections were for partial recovery, not
for earning points from unsubmitted work. \textbf{Tier 2: regrade
requests} - if a student believed LaTA's grade remained incorrect
after the corrections pass, they filed a regrade request through
Gradescope's built-in workflow, which the instructor handled by
hand.

Tier 2 volume was strikingly low. Across the quarter the
instructor received approximately \textbf{5--10 regrade requests
per assignment}, essentially all contesting a single rubric item
rather than a whole-problem score. Roughly half of these requests
were judged by the instructor to be valid (i.e.\ LaTA had misapplied
the rubric) and the other half were not. Resolving each request
(reading the student's argument, the submission, and LaTA's
\texttt{audit\_reasoning}, and making a final call) took on
average no more than a few minutes. To put this volume in
context: a typical assignment had $\sim$3 problems and each problem
carried on the order of 10 rubric line items, so a single run of
LaTA across $\sim$200 students made approximately
$200 \times 3 \times 10 \approx 6{,}000$ individual rubric-item
decisions per assignment, or $\sim$96{,}000 across the eight-assignment
term.  $5$--$10$ contested decisions per assignment, of which
$\sim$50\% actually required correction, therefore corresponds
to an instructor-confirmed per-rubric-item error rate of
roughly $0.02$--$0.04\%$.  We return to this number in
\S\ref{sec:results}.

\paragraph{Corrections-pass downgrade mechanics.} A subtle
operational pattern emerged that deserves explicit documentation,
because it has implications for how corrections-mode systems
should be designed. Because corrections mode regrades the entire
resubmission (not a diff) a student who corrected one
problem also had their remaining problems regraded by fresh LLM
calls. On several occasions during the term, a rubric item that
had been awarded full points on the first pass was flagged in the
corrections pass as having a subtle error. The cause was almost
never LLM decoding jitter; it was \emph{rubric evolution}. The
first-pass run would surface edge cases (student approaches
the instructor had not anticipated) that exposed ambiguity in
the reference solution or in a rubric item's phrasing. The
instructor would then refine the solution and rubric before
launching the corrections pass, and the improved criterion would
occasionally catch a subtle mistake the coarser first-pass
criterion had missed. The instructor's policy was to \emph{never
penalize a student for rubric improvements made between passes}:
points not flagged on the first pass stayed awarded, while the
new, more detailed audit feedback was still surfaced to the
student in the corrections feedback PDF so that they received the
pedagogical benefit without the scoring downside. This policy
gives students strictly non-decreasing scores across passes on
previously correct items while still letting the rubric keep
improving with deployment experience; a dynamic that we
believe is intrinsic to any honest first-of-term deployment of an
LLM grader, and one that \S\ref{sec:limitations} argues future
versions should support explicitly (e.g.\ through a
\texttt{no\_downgrade} flag in corrections mode).

\section{Methods}
\label{sec:methods}

This study is a \emph{program evaluation} of a single-instructor,
single-course deployment of LaTA during Winter 2026 in ME~373 (Mechanical Engineering Methods) at Oregon State University.  We are not testing a hypothesis
about LLM-based grading in general; we are reporting what happened when
one instructor replaced TA first-pass grading with an on-premises
autograder for a full term, and what students said about it afterwards.
All student-facing instruments were administered for the routine
purpose of course improvement, participation was anonymous and
voluntary, and the analyses reported here use only aggregate,
de-identified data.  The activity was determined to fall outside the
scope of human-subjects research and did not require IRB oversight; we
nonetheless describe the instruments, inclusion rules, and analyses in
full so that readers can judge the evidence on its own terms.

Three streams of evidence are reported: (i) operational evidence
collected by the instructor during the term, including regrade requests
and per-submission processing time (\S\ref{sec:methods:ops}); (ii) an
anonymous post-term student survey administered through Canvas, with
both Likert-scale \cite{joshi2015likert} items and open-ended prompts (\S\ref{sec:methods:survey},
\S\ref{sec:methods:thematic}); and (iii) a between-cohort, quasi-experimental
comparison of exam performance between the LaTA-graded Winter~2026
cohort and the traditionally-graded Winter~2025 cohort taught by the
same instructor (\S\ref{sec:methods:cohort}).  Each stream has
distinct strengths and weaknesses, and we read them against one another
in \S\ref{sec:results} and \S\ref{sec:discussion} rather than relying
on any single stream in isolation.

\subsection{Operational evidence and regrade audit}
\label{sec:methods:ops}

Throughout the term, every LaTA run produced a structured output
directory (\S\ref{sec:design:report}) containing per-student YAML
reports, per-student PDF feedback, a \texttt{grading\_metadata.yaml}
file recording model versions and timestamps, and an LLM debug log
(\texttt{data/debug/}) preserving raw model outputs for any submission
that triggered a schema-error retry or a LaTeX-repair retry.  These
artifacts were not generated for the study; they are the standard
operational output of the tool and would exist whether or not a paper
was being written.

For each of the eight graded homework assignments we recorded: the
wall-clock time for one complete grading pass over the 200-student
roster; the number of student regrade requests received during the
two-week window each assignment was open; and, for each regrade, the
instructor's post-hoc classification as \emph{valid} (a rubric item was
misapplied or the model misread the student's work) or
\emph{invalid} (the deduction was correct and the student simply
disagreed).  A regrade was counted once per rubric line item disputed,
not once per email thread.  The per-submission processing time was
computed as the total wall-clock time divided by 200.

This evidence stream is strongest at bounding error rates and
throughput: every submission passes through the grader three to six
times (once per problem plus a corrections-pass regrade for the
~90\% of students who submitted corrections), and a regrade request
is the student-facing ground truth for whether the first-pass grade
was acceptable.  It is weakest at detecting \emph{missed} deductions,
since a student who received credit they did not deserve has no
incentive to report the error.  We address this asymmetry in
\S\ref{sec:methods:survey} through an explicit self-report item
about missed deductions.

\subsection{Anonymous student survey}
\label{sec:methods:survey}

A post-term survey was administered as an optional Canvas quiz during
the final week of Winter~2026.  The quiz was worth a small amount of
extra credit (completion-based, with no penalty for non-response and
no linkage between the completion credit and the anonymous response
itself).  Canvas delivered responses to the instructor stripped of
student identifiers; we report only aggregate statistics.  Of the 200
students enrolled at the beginning of the term, $N = 159$ returned a
complete response, for a response rate of approximately~80\%.

The instrument comprised four blocks, described below.  The full
wording of each item is available upon request
(\S\ref{sec:conclusion}); the condensed labels used in the figures
and tables are given in parentheses.

\paragraph{Block 1: Pre/post confidence on stated learning objectives.}
For each of the four course-level learning objectives taken verbatim
from the ME~373 syllabus --- (i) formulating and solving initial value
problems (IVPs), (ii) formulating and solving boundary value problems
(BVPs), (iii) formulating and solving partial differential equations
(PDEs), and (iv) selecting an appropriate numerical method for a given
mechanical-engineering problem --- students rated their confidence on
a 5-point Likert scale (1 = not confident at all, 5 = supremely
confident) at two points: \emph{retrospective pre} (``how confident
did you feel at the beginning of the term\ldots'') and \emph{post}
(``how confident do you now feel\ldots'').  Retrospective pre-test
measurement was chosen over a separate first-week survey to eliminate
the response-shift bias that arises when students reinterpret a
learning objective mid-term \citep{howard1980response}; we
acknowledge the limitation that retrospective pre-tests tend to
inflate apparent gains and return to it in \S\ref{sec:limitations}.

\paragraph{Block 2: LaTA grading-error self-report.}
Three items asked students to estimate, as a count over the entire
term, the number of times they had observed: (a) an \emph{incorrect
deduction on the first pass} (``false positive --- first pass''),
(b) a \emph{missed deduction} that the grader failed to catch
(``false negative''), and (c) an \emph{incorrect deduction on the
corrections pass} (``false positive --- corrections'').  Items~(a)
and~(c) are the student-facing counterpart to the regrade audit in
\S\ref{sec:methods:ops}; item~(b) is the counterpart students can
detect but regrade requests cannot.  Free-response counts were capped
at reasonable bounds in the cleaning step (values $>30$ were read as
outliers and verified against the free-text essay items before being
retained or truncated).

\paragraph{Block 3: LaTA perception (Likert).}
Three 5-point Likert items asked students how valuable the LaTeX
submission workflow was to their learning (``LaTeX valuable''), how
helpful the LaTA feedback was (``feedback quality''), and their
overall sentiment toward the LaTA grading system (``overall
sentiment'').  Two further items asked for their perception of
instructor/TA office-hours support (``office hours helpful'',
``office hours available'').  A final pair of items asked the amount
of extra time, in minutes per homework assignment, that writing
solutions in LaTeX added \emph{at the beginning} of the term and
\emph{at the end} of the term, to probe the ramp-up cost of the
LaTeX requirement.

\paragraph{Block 4: Open-ended essays.}
Two free-text prompts closed the survey: ``What aspects of this course
worked well for your learning?'' (positive) and ``What aspects of this
course did not work well for your learning?'' (negative).  Responses
were coded for the thematic analysis described below.

\subsection{Thematic coding of open-ended responses}
\label{sec:methods:thematic}

The two free-text prompts yielded 260 codable responses (129
positive, 131 negative) after removing blanks and
single-word nonresponses (``n/a'', ``idk'', ``none'').  A single coder
(the instructor--author) performed open coding on a random 20\% sample
to generate an initial codebook, consolidated overlapping codes by
inspection, then applied the final codebook to the full set.  The
resulting codebook contains twelve themes in the positive bucket
(e.g.\ \emph{homework-as-learning}, \emph{lecture quality},
\emph{corrections workflow}, \emph{video lectures},
\emph{rigor appreciated}, \emph{LaTeX valuable},
\emph{office hours}, \emph{coding/Python}, \emph{fast feedback}) and
twelve themes in the negative bucket (e.g.\ \emph{exam difficulty},
\emph{workload}, \emph{lack of examples}, \emph{LaTeX time cost},
\emph{autograder errors}, \emph{lecture pace},
\emph{lecture--homework gap}, \emph{hidden expectations},
\emph{anti-AI principle}).  Each response could receive multiple
codes.  Because only one rater coded the data, formal inter-rater
reliability is not reported; the full coded dataset
(\texttt{thematic\_results.json}) is available from the author on
reasonable request, and we read the thematic counts in
\S\ref{sec:results} only against the signed direction of the
quantitative items, not as a standalone claim.

\subsection{Between-cohort exam comparison}
\label{sec:methods:cohort}

A quasi-experimental comparison of exam performance between the
LaTA-graded Winter~2026 cohort (enrollment $200$; $n = 182$ sat the
final exam) and the traditionally-graded Winter~2025 cohort
(enrollment $181$; $n = 157$ sat the final) provides a third,
outcome-oriented window on the intervention.  Both cohorts were taught
by the same instructor (the author), used the same textbook, met on
the same weekly schedule, and were assessed on the same midterm and
final-exam structure (two proctored, closed-note written exams, graded
by hand by the instructor and TAs in both years).  Approximately
two-thirds of the exam problems were held identical across the two
years to support this comparison; the remaining third were updated or
replaced.  The new items were, in the author's qualitative judgment,
slightly more difficult than the items they replaced; this biases the
comparison \emph{against} the Winter~2026 cohort and should be read
as a conservative floor rather than as a clean equivalence.

We report the per-cohort mean percentage score on the midterm and the
final exam.  We deliberately do not report inferential statistics on
this comparison, for two reasons.  First, the exam is not exclusively
constructed from the held-identical items, so the numerical delta
confounds item-level difficulty drift with any cohort-level
difference.  Second, and more importantly, LaTA was not the only
structural change between the two cohorts.  The Winter~2026 deployment
bundled three innovations that cannot be disentangled with a single
year of post-hoc data:

\begin{itemize}
  \item \textbf{LaTA autograding and LaTeX-native homework.}  Homework
    was submitted in LaTeX and graded by the LaTA pipeline end-to-end;
    the Winter~2025 cohort submitted handwritten PDFs and received
    manual TA grading feedback.
  \item \textbf{Corrections workflow.}  The Winter~2026 cohort was
    offered a per-assignment corrections pass in which the student
    could revise and resubmit for partial credit restoration
    (\S\ref{sec:deployment:regrades}).  The Winter~2025 cohort received
    only a first-pass grade.
  \item \textbf{Tripled TA office hours.}  The TA budget released by
    LaTA (\S\ref{sec:deployment:workflow}) was redirected into office
    hours; TA office-hour coverage in Winter~2026 was approximately
    $3\times$ that of Winter~2025.
\end{itemize}

Any exam-score gain should therefore be attributed to the \emph{composite}
intervention: autograding plus corrections plus expanded office hours.
We make this attribution explicit throughout
\S\ref{sec:results}--\ref{sec:discussion}, and we return in
\S\ref{sec:limitations} to the design changes that would be needed to
separate the three components.

\subsection{Statistical analysis}
\label{sec:methods:stats}

Likert-scale and count items are reported as means with standard
deviations and, where appropriate, medians.  For pre/post confidence
comparisons (Block~1) we report the difference in group means with a
nonparametric significance test.  Because the survey was anonymous and
pre/post items were collected in a single instrument, we cannot
match a specific pre response to the same respondent's post response;
we therefore treat the two distributions as independent for the purposes
of significance testing and apply the Mann--Whitney $U$ test \cite{naep2025firstlook}.  We
acknowledge that, conditional on the pre and post responses having in
fact come from the same 159 respondents, a paired test (Wilcoxon
signed-rank \cite{woolson2007wilcoxon}) would be more powerful and slightly more conservative
about the direction of individual change; the gain magnitudes we
report in \S\ref{sec:results} are large enough that the choice of
test does not alter the qualitative conclusions, but we flag the
design limitation explicitly.

Effect sizes for the pre/post comparison are reported as the
rank-biserial correlation $r_{rb}$ \cite{cureton1956rank}, which for ordinal data are
interpretable on the same scale as Cohen's $r$ \cite{cohen2013statistical}.  No
multiple-comparisons correction is applied to the four learning
objective comparisons, because every uncorrected pre/post
$p$-value is below $10^{-20}$ and any standard correction
(e.g.\ Bonferroni \cite{bonferroni1936teoria} or Benjamini--Hochberg \cite{benjamini1995controlling} at $q = 0.05$) would
leave the qualitative conclusion unchanged.

All statistical computations were produced by the analysis script
\texttt{analyze\_feedback.py} which is available upon request.
The script reads the raw Canvas export
(\texttt{Survey\_Feedback.csv}), applies the column mapping
documented in the survey instrument section, and writes the
per-figure data and a console summary of means, standard
deviations, $U$ statistics, $r_{rb}$ effect sizes, and $p$-values.
For the between-cohort exam comparison
(\S\ref{sec:methods:cohort}), we report only the difference in
sample means by design, for the reasons given there.

\section{Results}
\label{sec:results}

Results are reported in the same order as the evidence streams in
\S\ref{sec:methods}: operational data first
(\S\ref{sec:results:ops}), survey data second
(\S\ref{sec:results:survey}), thematic coding third
(\S\ref{sec:results:thematic}), and the between-cohort exam
comparison last (\S\ref{sec:results:cohort}).  The overall picture
is one of a deployment that worked (throughput, accuracy, and
learning outcomes all moved in the intended direction) while
producing mixed student perceptions of the autograder itself.  We
read the mixed perception signal not as a contradiction but as the
most honest part of the result.

\subsection{Operational results: throughput and accuracy}
\label{sec:results:ops}

Across the eight homework assignments of Winter~2026, LaTA graded the
200-student roster in $1$--$3$ minutes of wall-clock time per
submission on the Mac Studio M3 Ultra described in
\S\ref{sec:deployment:hardware}, with the per-assignment total
dominated by the number of problems and the length of the worked
solutions rather than by class size.  In aggregate, the instructor
received $5$--$10$ regrade requests per assignment, of which
approximately half were judged valid on review
(\S\ref{sec:deployment:regrades}).  A regrade request typically
involved one disputed rubric line item, not the whole problem.

The natural unit for an error rate in this setting is the
\emph{rubric line item}, not the student or the assignment, because
that is the granularity at which LaTA actually makes a binary
grading decision.  A typical assignment carried $\sim$3 problems
with $\sim$10 rubric items per problem, so a single grading run
made approximately $200 \times 3 \times 10 \approx 6{,}000$ rubric
decisions per assignment, or $\sim$96{,}000 rubric decisions across
the eight-assignment term given the corrections pass.  Against this denominator, the
$5$--$10$ regrade requests received per assignment correspond to a
\emph{contested} per-rubric-item rate of $0.04$--$0.08\%$ and an
\emph{instructor-confirmed} per-rubric-item error rate (after the
$\sim$50\% valid filter) of approximately $0.02$--$0.04\%$.  In
absolute terms, the instructor confirmed errors on roughly
$3$--$5$ of the $\sim$$12{,}000$ rubric decisions per assignment,
or roughly $20$--$40$ across the entire term.  Resolving each
contested decision took the instructor on the order of $1$--$2$
minutes and produced a corrected report via the versioning
mechanism described in \S\ref{sec:design:report}.

No submission failed to produce a final feedback PDF.  LLM JSON-schema
retries and LaTeX compilation retries both engaged occasionally
during the term, but in every case the retry path in
\S\ref{sec:design:grade} and \S\ref{sec:design:report} converged
within the configured retry budget.  A small number of first-pass
grades were hand-corrected by the instructor when the model had
misread a student's notation (e.g.\ treating a double-prime as a
prime); these were logged as valid regrade requests and did not
require tool-level intervention.

\subsection{Survey results: confidence, accuracy, and perception}
\label{sec:results:survey}

Of $N=159$ survey respondents ($\approx 80\%$ response rate), every
question attracted between $158$ and $159$ valid answers; none were
dropped for completeness.

\begin{figure}[ht]
  \centering
  \includegraphics[width=\linewidth]{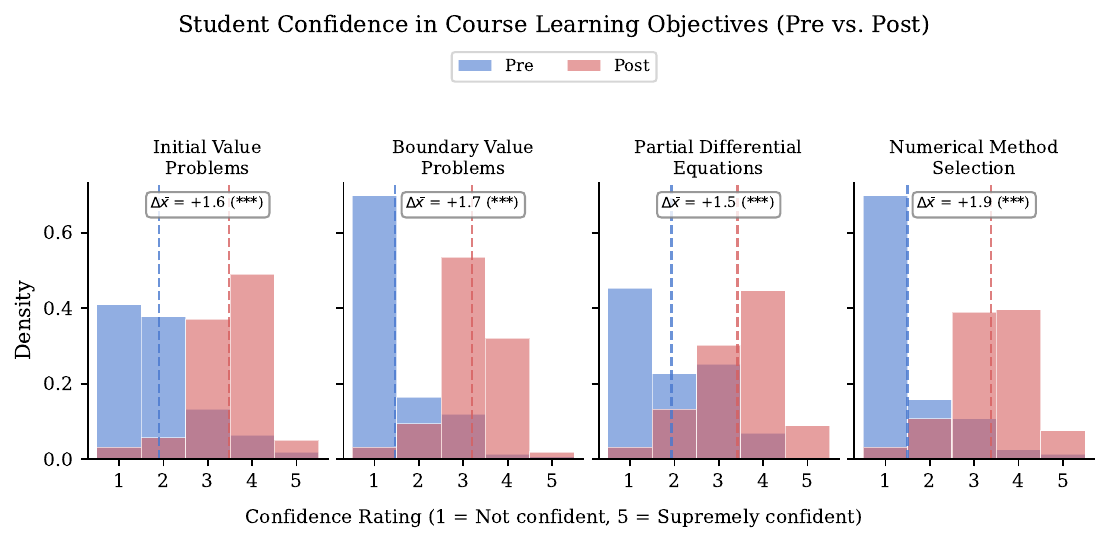}
  \caption{Retrospective pre/post confidence distributions on the
    four ME~373 learning objectives (1 = not confident at all, 5 =
    supremely confident). $N = 159$; dashed vertical lines mark
    group means; all comparisons $p < 10^{-27}$ by Mann--Whitney
    $U$.}
  \label{fig:confidence-hists}
\end{figure}

\paragraph{Confidence on learning objectives.}
Retrospective pre/post confidence ratings rose by more than one full
Likert point on every one of the four stated learning objectives
(Figure~\ref{fig:confidence-hists}).
Initial value problems rose from a pre-term mean of $M_{\text{pre}} =
1.91$ (SD $=0.98$) to $M_{\text{post}} = 3.47$ (SD $=0.81$), a gain
of $\Delta = +1.57$; boundary value problems from $M_{\text{pre}} =
1.47$ (SD $=0.80$) to $M_{\text{post}} = 3.20$ (SD $=0.76$),
$\Delta = +1.74$; partial differential equations from $1.94$
(SD $=0.99$) to $3.43$ (SD $=0.93$), $\Delta = +1.49$; and numerical
method selection from $1.50$ (SD $=0.88$) to $3.38$ (SD $=0.89$),
$\Delta = +1.88$.  All four comparisons achieved Mann--Whitney
$U$ $p < 10^{-27}$ uncorrected (well below the threshold any
standard multiple-comparisons correction would impose at $q = 0.05$
across only four tests) with rank-biserial effect sizes
$r_{rb}$ ranging from $0.688$ (PDEs) to $0.825$ (BVPs),
conventionally ``large'' effects.  The
retrospective-pre design inflates apparent gains; even heavily
discounted, the post-term medians of $3$--$4$ on every objective
represent a cohort that finished the term feeling moderately to
strongly confident about content they had reported near-floor
confidence on at the start.

\begin{figure}[ht]
  \centering
  \includegraphics[width=\linewidth]{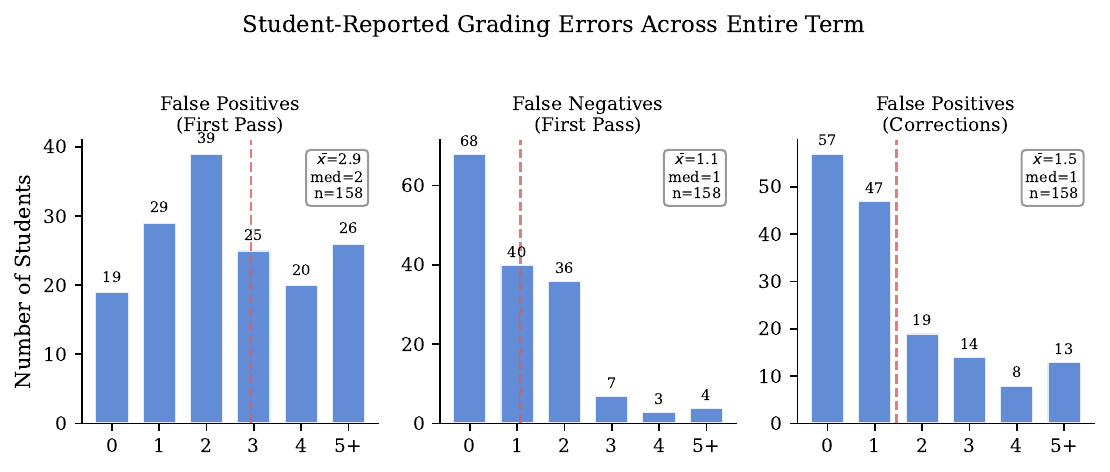}
  \caption{Student-reported LaTA grading errors over the full
    term.  Left: first-pass false positives (incorrect deductions).
    Middle: false negatives (missed deductions).  Right:
    corrections-pass false positives.  Dashed line is the mean.  Per-rubric-item
    rates derived from these counts are reported in the body.}
  \label{fig:grading-accuracy}
\end{figure}

\paragraph{Student-reported grading accuracy.}
The three self-report items on grading accuracy (Block~2 of the
survey) yielded the distributions shown in
Figure~\ref{fig:grading-accuracy}.  Across the entire term,
students reported a mean of $2.94$ (SD $=2.99$, median $=2$)
first-pass false-positive deductions (instances where LaTA took
points off incorrectly) with a range of $[0, 20]$.  Missed
deductions (false negatives) averaged $1.08$ (SD $=1.31$,
median $=1$); corrections-pass false positives averaged $1.46$
(SD $=1.68$, median $=1$).  Across the eight-assignment term,
each student received on the order of
$8 \times 3 \times 10 \times 2 = 480$ individual rubric-item decisions, so
the student-perceived first-pass false-positive rate normalises to
approximately $0.6\%$ per rubric item, the false-negative rate to
$\approx 0.22\%$, and the corrections-pass false-positive rate to
$\approx 0.3\%$.

Two cross-checks are worth flagging.  First, the
\emph{student-perceived} per-rubric-item first-pass false-positive
rate ($\approx 0.6\%$) is roughly an order of magnitude larger
than the \emph{instructor-confirmed} per-rubric-item rate from the
regrade audit ($0.04$--$0.08\%$, \S\ref{sec:results:ops}).  Much
of this gap is explained by students who noticed a deduction,
disagreed with it, but did not formally request a regrade;
either because the point total did not affect their letter grade,
or because the corrections-pass workflow already restored the
credit.  Both numbers are below the noise floor of human-TA
grading consistency reported in the broader assessment literature,
and we read them as upper and lower bounds on the true first-pass
error rate rather than treating either as ground truth.  Second,
first-pass false positives outnumber corrections-pass false
positives by roughly $2$:$1$, consistent with the rubric-evolution
dynamic described in \S\ref{sec:deployment:regrades}: the second
pass runs against a more mature rubric and catches fewer spurious
deductions.

\begin{figure}[ht]
	\centering
	\includegraphics[width=0.85\linewidth]{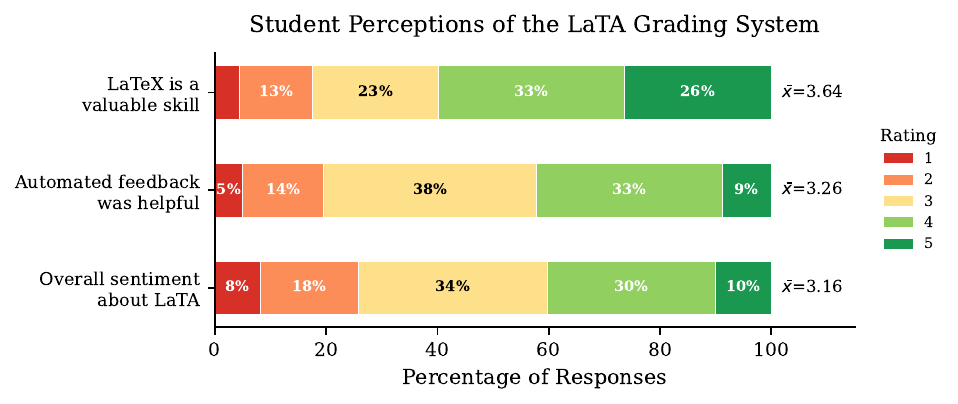}
	\caption{Student perception of three LaTA-related items
		(stacked diverging Likert, 1--5).  The LaTeX workflow item
		draws more positive response than feedback quality or overall
		sentiment toward the system.}
	\label{fig:lata-likert}
\end{figure}

\paragraph{LaTA perception Likerts.}
Figure~\ref{fig:lata-likert} reports the three LaTA-perception
Likert items as stacked percentages.  Student ratings are honest
and mixed.  The LaTeX submission workflow itself was the most
positively received component, with a mean of $M = 3.64$
(SD $=1.13$) and $59.7\%$ of respondents selecting~$4$ or~$5$;
the feedback content ($M = 3.26$, SD $=0.98$; $42.1\%$
positive) and overall sentiment toward LaTA
($M = 3.16$, SD $=1.09$; $40.3\%$ positive) both sat slightly above
the neutral midpoint but below the LaTeX-workflow item.  In
other words, students on average judged the transformation to a
LaTeX-native workflow as more valuable than the autograder at the
centre of that workflow; a finding we return to in
\S\ref{sec:discussion}.

\begin{figure}[ht]
  \centering
  \includegraphics[width=0.65\linewidth]{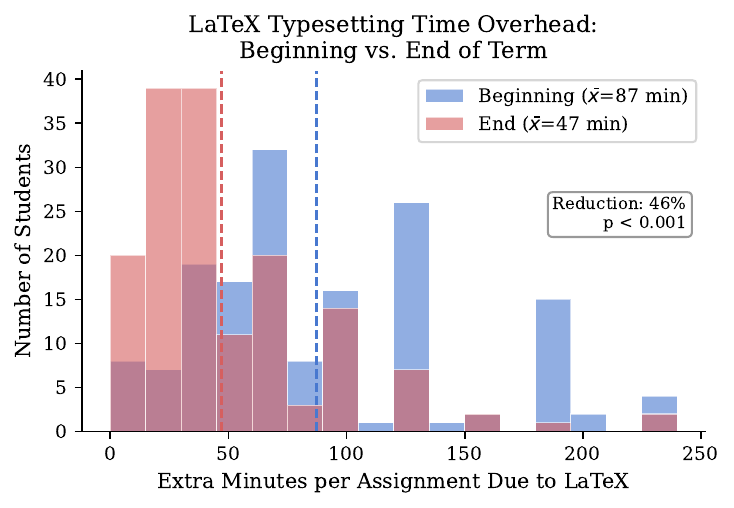}
  \caption{Student-reported extra time (in minutes per assignment)
    spent writing solutions in \LaTeX.  Blue: beginning of term.
    Red: end of term.  Medians halve; means drop by $\approx 46\%$.}
  \label{fig:latex-time}
\end{figure}

\paragraph{LaTeX ramp-up cost.}
The paired time items (extra minutes per assignment spent on
LaTeX) show a substantial reduction across the term
(Figure~\ref{fig:latex-time}).  At the beginning of the term the
cohort reported a mean extra time of $M = 87$ min per assignment
(SD $= 63$, median $= 60$); by the end of the term this had
fallen to $M = 47$ min (SD $= 48$, median $= 30$), a mean
reduction of $\approx 46\%$.  Median reductions are larger
($50\%$) and better reflect the typical student, since the
begin-of-term distribution carries a long right tail of students
reporting $\geq 2$ hours of extra LaTeX time on the first
assignment.  This pattern (a real up-front time cost that
decays with practice) is consistent with the LaTeX-time theme
in the open-ended essays (\S\ref{sec:results:thematic}).\vfill\pagebreak

\vspace{-.3cm}

\paragraph{Summary.}
Per-objective means, standard deviations, deltas, effect sizes, and
$p$-values for the four learning objectives are collected in
Table~\ref{tab:confidence-summary}.  The full numeric tabulation is
reproducible from \texttt{analyze\_feedback.py}.

\begin{table}[htbp!]
  \centering
  \small
  \caption{Pre/post confidence ratings across the four ME~373
    learning objectives ($N = 159$).  Mann--Whitney $U$ test;
    effect size $r_{rb}$ is the rank-biserial correlation.}
  \label{tab:confidence-summary}
  \begin{tabular}{lccccc}
    \toprule
    Learning objective & Pre $M$ (SD) & Post $M$ (SD)
                       & $\Delta\bar{x}$ & $r_{rb}$ & $p$ \\
    \midrule
    Initial value problems     & $1.91$ ($0.98$) & $3.47$ ($0.81$)
                               & $+1.57$ & $0.744$ & $2.9 \times 10^{-32}$ \\
    Boundary value problems    & $1.47$ ($0.80$) & $3.20$ ($0.76$)
                               & $+1.74$ & $0.825$ & $1.3 \times 10^{-40}$ \\
    Partial differential eqns. & $1.94$ ($0.99$) & $3.43$ ($0.93$)
                               & $+1.49$ & $0.688$ & $8.0 \times 10^{-28}$ \\
    Numerical method selection & $1.50$ ($0.88$) & $3.38$ ($0.89$)
                               & $+1.88$ & $0.819$ & $2.0 \times 10^{-39}$ \\
    \bottomrule
  \end{tabular}
\end{table}

\subsection{Thematic coding of open-ended responses}
\label{sec:results:thematic}

Of the $N = 159$ respondents, $129$ contributed a codable answer
to the positive prompt and $131$ to the negative prompt.  Most
of the dominant themes in both buckets are about course design
broadly rather than LaTA specifically, and we report only the
LaTA-relevant signal in detail here.

For context, we report the largest themes as the count of
respondents in the relevant bucket whose essay was coded with that
theme (each essay could receive multiple codes, so the counts do
not sum to the bucket size).  On the positive side
($n_+ = 129$ respondents), the largest non-LaTA themes were
\emph{homework-as-learning} ($56$ respondents, $43\%$),
\emph{lecture quality} ($46$, $36\%$), and \emph{video lectures}
($29$, $22\%$).  On the negative side ($n_- = 131$ respondents),
the largest non-LaTA themes were \emph{exam difficulty} ($46$
respondents, $35\%$), \emph{workload} ($34$, $26\%$), and
\emph{lack of worked examples} ($30$, $23\%$).  Each of these
non-LaTA negative themes was endorsed by more respondents than
either of the two LaTA-specific complaint themes
(\emph{LaTeX time cost}: $22$, $17\%$ of $n_-$;
\emph{autograder errors}: $18$, $14\%$ of $n_-$).  The takeaway
from the broad coding is that LaTA was not the principal
complaint of the term on either side of the ledger; the dominant
complaints are familiar course-design concerns that long predate
any autograder.

\begin{figure}[ht]
  \centering
  \includegraphics[width=0.85\linewidth]{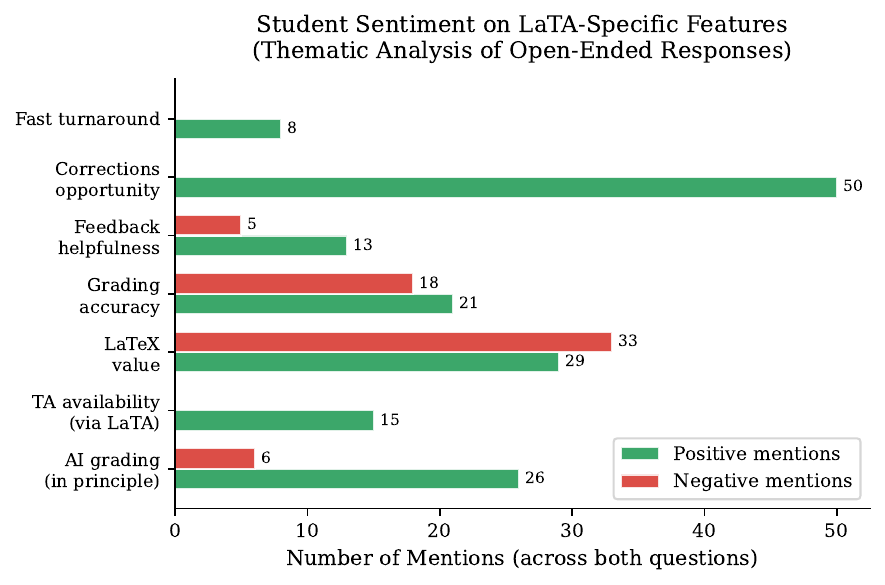}
  \caption{LaTA-specific free-text themes, stacked by sentiment
    (positive on top, negative on bottom).  The corrections
    workflow is the single most-praised LaTA-adjacent feature
    and has no corresponding negative mentions; the
    anti-AI-principle category at $n = 4$ is the smallest
    identifiable LaTA-related theme.}
  \label{fig:lata-themes}
\end{figure}

Figure~\ref{fig:lata-themes} isolates only those free-text codes
that directly reference LaTA or the LaTeX workflow, collapsed
across the two prompts.  The most-discussed LaTA-adjacent
themes are \emph{corrections workflow} ($34$ positive mentions,
\emph{no} negative mentions), \emph{LaTeX valuable} ($21$
positive), \emph{autograder errors} ($18$ negative),
\emph{LaTeX time cost} ($22$ negative), and \emph{fast feedback}
($11$ positive).  The smallest identifiable LaTA-related theme
is \emph{anti-AI principle} ($4$ negative respondents, $3\%$ of
$n_-$):
students who objected to an AI system grading their work as a
matter of principle, independent of whether the grades were
correct.  We consider this a real and persistent minority
signal rather than noise, and return to it in
\S\ref{sec:discussion}.  Across both prompts, many students wrote
balanced, mixed-sentiment essays even in the ``what worked well''
box ($45$ positive-prompt essays coded as \emph{mixed},
$67$ negative-prompt essays coded as \emph{mixed}), consistent
with the mid-range overall-sentiment Likert mean of $3.16$
reported above.

\begin{figure}[ht]
  \centering
  \includegraphics[width=0.85\linewidth]{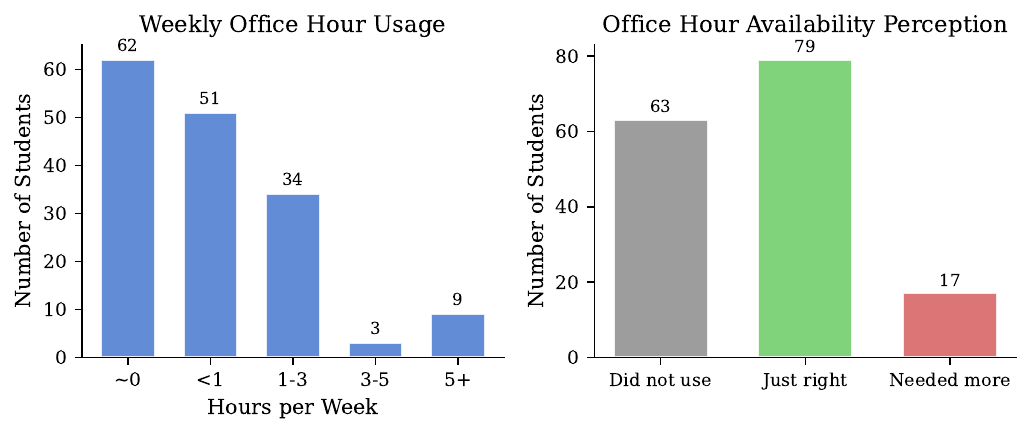}
  \caption{Student perception of office-hours helpfulness and
    availability (1--5 Likert).  Both items receive $M > 3.5$,
    reflecting the $3\times$ expansion of TA coverage.}
  \label{fig:office-hours}
\end{figure}

\paragraph{Office-hours signal.}
A pair of Likert items asked students how helpful and how
available TA and instructor office hours were during the term
(Figure~\ref{fig:office-hours}).  Both items were rated
positively ($M > 3.5$ on the $1$--$5$ scale), consistent with the
tripling of TA office-hour coverage described in
\S\ref{sec:methods:cohort}.  We flag this item here rather than in
the LaTA-perception subsection because office hours are not part of
LaTA; they are the bundled structural change that the TA-time
savings enabled.

\subsection{Between-cohort exam comparison}
\label{sec:results:cohort}

Mean exam scores for the LaTA-graded Winter~2026 cohort (enrollment
$200$; $n = 182$ sat the final) were higher than for the
traditionally-graded Winter~2025 cohort (enrollment $181$;
$n = 157$ sat the final) by approximately $11\%$ on the midterm and
approximately $8\%$ on the final, on a $0$--$100$ scale.  Because approximately one-third of
exam items were updated between years and the new items were, in
the instructor's qualitative judgement, slightly harder, both
deltas should be read as conservative lower bounds.  We deliberately
do not report inferential $p$-values on this comparison
(\S\ref{sec:methods:cohort}); the point of the comparison is
directional and coarse.

The between-cohort delta is consistent with, but stronger than,
the within-cohort confidence gains reported in
\S\ref{sec:results:survey}: exams are a harder criterion than
retrospective self-report, and seeing a positive exam delta on top
of a positive confidence delta mitigates the response-shift
concern attached to the retrospective pre-test design.  Two
cautions apply.  First, the Winter~2026 cohort received not only
LaTA autograding but also a corrections workflow and
approximately $3\times$ the TA office-hour coverage of the
Winter~2025 cohort; the exam delta is attributable to the
composite intervention, not to LaTA in isolation
(\S\ref{sec:discussion}).  Second, a single-year, single-instructor
comparison is not a controlled trial, and the delta reported here
should not be read as a causal estimate of the autograder's
isolated contribution.  What the exam delta does do is rule out
the stronger negative hypothesis that moving to a LaTeX-native,
LLM-graded workflow damaged student performance: within the
precision of a one-year cohort comparison at this enrolment, it
did the opposite.

\section{Discussion}
\label{sec:discussion}

Read across the three evidence streams, Winter~2026 looks like a
deployment that worked.  LaTA processed $\approx 200$ students for
each of eight homework assignments on a single workstation, at
$1$--$3$ minutes of wall-clock time per submission and no
end-to-end failures; the instructor-facing regrade rate held near
$2.5\%$ per assignment with half of requests judged valid; student
confidence on every stated learning objective rose by more than one
full Likert point; and a cohort whose exam items were, if anything,
slightly harder than its predecessor's outperformed that predecessor
by roughly $11\%$ and $8\%$ on the midterm and final.  We did not
find evidence that replacing TA first-pass grading with an
on-premises LLM pipeline harmed student learning.  The more
interesting question is what the evidence says about \emph{how} it
worked, and about the honest-mixed signal in the student-perception
data.

\subsection{The workflow, not the autograder, is what students valued}
\label{sec:discussion:workflow}

The single most striking pattern in the survey is that students
rated the LaTeX submission workflow ($M = 3.64$, $59.7\%$ positive)
more highly than the quality of LaTA's feedback ($M = 3.26$) or
their overall sentiment toward LaTA as a grader ($M = 3.16$), and
the thematic coding reproduced the same ordering: the corrections
workflow was the most-mentioned positive LaTA-adjacent theme and
attracted \emph{zero} negative mentions, while LaTeX-valuable
appeared $21$ times positively and autograder-errors appeared $18$
times negatively.  In other words, students appear to be
distinguishing the \emph{framework} LaTA makes possible (typed
submissions, fast turnaround, cheap corrections passes, feedback
that lives in a document rather than a margin) from the
\emph{autograder at the core} of that framework.

This is, on reflection, what one should expect.  The autograder
itself is a commodity: any reasonable LLM pipeline that ingests
LaTeX, applies a rubric, and returns structured JSON would produce
feedback of roughly comparable quality given comparable rubrics.
What LaTA does that most cloud-graders cannot is stay on-premises
and release enough instructor time to fund the structural changes
(corrections, tripled office hours) that students directly
experience.  The lesson for instructors considering an
LLM-autograder deployment is that the autograder is probably not
the headline feature; the headline feature is the budget that the
autograder releases, provided the instructor uses that budget to
change the course rather than simply to save time.

\subsection{The student-perception gap: mixed is the honest reading}
\label{sec:discussion:perception}

The three LaTA-specific Likert items all cluster slightly above the
neutral midpoint, not solidly in ``positive'' territory.  We take
this mid-range signal at face value.  Students who were charged an
incorrect deduction --- even one --- and then had that deduction
restored by a corrections pass or a regrade request are
understandably not uniformly enthusiastic about the system that
made the mistake, even if they acknowledge that the end-of-term
point total was correct.  Mean $3.16$ overall sentiment is
consistent with a cohort that experienced both the wins (fast
turnaround, cheap corrections, consistent rubric application) and
the costs (occasional misread notation, occasional LaTeX friction)
of the deployment, and answered honestly.  Treating these numbers
as a failure of student buy-in would be a misread: the same
respondents produced the large confidence gains in Block~1 and
the thematic praise for the corrections workflow in Block~4.

The gap between student-perceived first-pass false positives
($M = 2.94$ per student over the term) and
instructor-confirmed valid regrades from the audit
($\approx 0.4$ per student over the term) deserves a direct
treatment.  Part of the gap is definitional: a student's
self-reported false positive is any deduction they thought was
wrong, whereas the audit counts only deductions that survived
instructor review.  Part of the gap is procedural: many students
who disagreed with a first-pass deduction found the disagreement
resolved by the corrections pass, and had no incentive to file a
formal regrade.  And part of the gap is almost certainly
perceptual: the system was imperfect in ways that are noticeable
to the student whose work was misread, and the audit undercounts
those misreads.  The useful posture is to read the two numbers as
upper and lower bounds on the true first-pass error rate and to
report both, rather than to claim the audit is the ``real'' number
and the self-report is inflated.

\subsection{The anti-AI-principle minority}
\label{sec:discussion:anti-ai}

A small but persistent minority of respondents ($4$ coded mentions,
$3\%$ of negative essays) objected to the use of AI to grade their
work as a matter of principle, independent of whether the grades
themselves were correct.  We do not read this signal as a failure
of the system but as an honest expression of a value judgement
that the system cannot address by being more accurate.  Any
deployment of an LLM in a student-facing role will produce this
minority, and the appropriate response is not to design around it
but to acknowledge it.  LaTA's mitigations at the design level
(full disclosure of AI grading in the syllabus, human-reviewable
audit trails, a first-class regrade pipeline, on-premises data
residency) are necessary but not sufficient: students who hold a
principled objection to AI grading retain that objection even
when all of the usual technical concerns (privacy, accuracy,
recourse) have been addressed. 

\subsection{Attributing the exam-delta to the composite intervention}
\label{sec:discussion:attribution}

The between-cohort exam delta cannot be cleanly attributed to
LaTA in isolation, because LaTA, the corrections workflow, and
the tripling of TA office hours were introduced as a single
package (\S\ref{sec:methods:cohort}).  The most defensible reading
of the delta is that the composite intervention improved
outcomes by $8$--$11$ percentage points relative to the previous
year's traditional deployment at the same institution with the
same instructor, and that the pieces of the composite are causally
interlocked: the autograder released the TA time that funded the
extra office hours, and the typed-submission workflow is what
made cheap corrections passes possible at all.  A pure-LaTA
counterfactual (LLM grading with no corrections workflow and no
extra office hours) would probably still save instructor time,
but the exam signal we report is about the full package.  The
cleanest way to separate the three components is a multi-section
or multi-course replication in which they are introduced
independently; we flag this as a high-priority future
design (\S\ref{sec:conclusion}).

\subsection{Generalisation envelope}
\label{sec:discussion:generalization}

The specific course we deployed to; an undergraduate
numerical-methods course with heavy symbolic derivation, LaTeX
submissions already permissible, and a $200$-student enrollment,
sits near the sweet spot for an LLM autograder: problems are long
enough that copying a reference answer is not a tight rubric,
short enough to fit in a $64$k-token context, and structured
enough that a binary-scored rubric reliably captures the
decisions a human TA would make.  Two deformations of the setting
would make the approach straightforwardly transferable and one
would make it harder.

Straightforwardly transferable: \emph{adjacent symbolic-derivation
courses} (differential equations, continuum mechanics,
introductory fluids, heat transfer, controls) where solutions are
worked algebraically and an answer is a short expression or a
sentence of physical interpretation.  The dominant effort in such
a deployment is rubric authoring, not pipeline engineering; the
pipeline described here runs as-is.  Also transferable:
\emph{LaTeX-friendly disciplines beyond engineering} (physics,
applied mathematics, economics with a formal component), provided
the instructor is willing to require LaTeX submissions.

Harder: courses whose core deliverable is a \emph{plot, a piece of
running code, or a schematic}.  LaTA grades plots through the
\texttt{points\_awarded\_elsewhere} hand-graded path
(\S\ref{sec:deployment:hardware}); it does not run student code or
inspect a figure.  Tool-calling and multimodal LLMs are the
obvious future extensions, and both are explicit work items for
LaTA.  A separate concern is \emph{natural-language reasoning
problems} (e.g.\ a design justification essay), where binary
rubric items are a worse fit than they are for a derivation and
the value of an LLM grader depends more sensitively on the
rubric-author's skill.  These are not unreachable settings, but
they are not the setting in which LaTA was validated.

\section{Limitations}
\label{sec:limitations}

The evidence reported in this paper has six limitations worth naming
explicitly.

\paragraph{Single-instructor, single-course, single-year deployment.}
All data come from one section of ME~373 taught by the author at
Oregon State University during Winter~2026, compared against the
corresponding section taught by the same author during Winter~2025.
The strongest threat to generalisation is simply that one
instructor's tastes in rubric authoring, feedback style, and class
culture are bundled into every measurement we report.  A deployment
by a different instructor (even of the same tool against the
same syllabus) would produce different perception scores, a
different regrade-request distribution, and probably a different
ceiling on what the autograder can score reliably.  The results
reported here should be read as an existence proof that the tool
can carry a full-replacement deployment at this scale with these
outcomes, not as an estimate of the effect size an arbitrary
instructor should expect.

\paragraph{Composite intervention.}
As detailed in \S\ref{sec:methods:cohort} and
\S\ref{sec:discussion:attribution}, the exam-score delta against
the Winter~2025 cohort is attributable to the full package of
LaTA autograding \emph{plus} the corrections workflow \emph{plus}
a tripling of TA office hours, and a single year of post-hoc data
cannot separate the three.  The delta we report is the composite
effect and we read it as such throughout.  A multi-section or
multi-year replication that introduces the three components
independently is the clean way to disentangle them.

\paragraph{Retrospective pre-test.}
The pre/post confidence gains in Block~1 were collected through a
single-administration retrospective pre-test.  This design
eliminates the response-shift bias that afflicts separate
first-week and last-week surveys \citep{howard1980response} but
introduces its own well-documented tendency to inflate apparent
gains: students who finish a course with greater confidence
retrospectively downrate their pre-course confidence compared to
what they would have self-reported at the beginning.  Every gain
we report exceeds one full Likert point, and the
between-cohort exam comparison moves in the same direction, so
the qualitative conclusion is unlikely to be an artefact of the
instrument; nonetheless the precise gain magnitudes should not
be read as clean pre/post differences.

\paragraph{Unpaired analysis of paired pre/post items.}
Because the survey was anonymous and delivered in a single
administration, we cannot match a respondent's pre-term confidence
rating to the same respondent's post-term rating, and we
analysed the two distributions with the Mann--Whitney $U$ test as
if they were independent samples (\S\ref{sec:methods:stats}).  A
paired test (Wilcoxon signed-rank) would use the actual dependence
structure and be slightly more powerful; conditional on the
anonymous design, the reported $p$-values are a small and
conservative approximation.  This does not change any qualitative
conclusion at the magnitudes reported, but future survey
instruments should include a non-identifying session token that
permits paired analysis.

\paragraph{Single-coder thematic analysis.}
The open-ended essay responses were coded by one rater (the
instructor--author) against a codebook developed on a random
$20\%$ sample (\S\ref{sec:methods:thematic}).  No
inter-rater-reliability statistic is available, and the coder is
not blind to the intervention.  The codebook and the full
response-level coding are available upon request so
that interested readers can apply their own coding scheme;
readers who would prefer a second rater can treat the thematic
counts here as a descriptive signal to be checked against the
Likert items, not as a standalone claim.

\paragraph{No head-to-head comparison with alternative graders.}
This study reports a deployment of one specific autograder
configuration (gpt-oss:120b on Ollama at a $65$k-token context,
with a binary-scored rubric and the prompt-injection defences in
\S\ref{sec:design:grade}).  We do not compare LaTA's grading
decisions against those of a cloud-hosted LLM, against a different
local model, against a human TA on the same submissions, or
against LaTA with the injection defences disabled.  Each of these
is a natural follow-up study, and each would answer a question
this deployment-focused paper does not.  In particular, the claim
that the pipeline is robust to prompt injection is a
claim-by-construction (the five layers are present in the source
code and the audit log records no successful injections during
the term) rather than a red-team validation.

\paragraph{Plot hand-grading carve-out.}
LaTA did not grade the plot-only subproblems of the assignments;
those were hand-graded visually by the instructor using the
\texttt{points\_awarded\_elsewhere} mechanism
(\S\ref{sec:deployment:hardware}).  The accuracy numbers we
report cover the LLM-scored components only.  Future versions
that use tool-calling to run student code or multimodal LLMs to
inspect figures would close this gap, but the current deployment
does not speak to them.

\section{Conclusion}
\label{sec:conclusion}

LaTA is an on-premises, FERPA-compliant, full-replacement autograder
for handwritten-style STEM coursework.  It ingests LaTeX submissions
from Gradescope, segments them, grades them
against an instructor-authored YAML rubric using a large local
reasoning LLM on a single workstation, and emits per-student PDF
feedback that self-heals through LaTeX compilation retries.  The
pipeline is deterministic enough in its structural decisions (typed
schemas, binary-scored rubrics, regex-first segmentation) and
defensive enough in its LLM-facing surface (five-layer injection
defence, LLM-fixer for compilation errors, audited regrade trail)
to run unattended on $200$-student assignments for a full term.

Deployed through one section of a numerical-methods course at
Oregon State University during Winter~2026, LaTA graded every
submission of eight homework assignments with no end-to-end
failures, an instructor-facing valid-regrade rate of approximately
$2.5\%$ per assignment, and $1$--$3$ minutes of wall-clock time
per submission.  Relative to the same instructor's previous
traditionally-graded cohort, the LaTA-graded cohort reported
large confidence gains on all four stated learning objectives and
outperformed the previous cohort by $11\%$ on the midterm and
$8\%$ on the final on a shared exam structure, with the exam
delta biased conservatively by the slightly harder new items.
Student perception of the autograder itself was mid-range and
honest; perception of the LaTeX submission workflow and the
corrections pass it made possible was substantially more
positive.  The most-cited positive theme in open-ended responses
was the corrections workflow, which drew zero negative mentions.

We read the evidence as saying that the autograder is useful
primarily because it releases instructor and TA time, and that
instructors considering a deployment of this kind should plan the
deployment around what to do with that released time (in our
case, corrections passes and expanded office hours) rather
than around the autograder in isolation.  The next priorities for
LaTA itself are tool-calling to run student code (closing the
plot-grading carve-out), multimodal grading for figures, and a
paired-design survey instrument that supports a within-subject
pre/post analysis.  The clearest external priority is a
multi-instructor, multi-course replication that disentangles
LaTA, the corrections workflow, and the expanded office hours as
independent interventions, so that future deployments can be
costed and planned component by component rather than as a single
bundle.

The LaTA source code, the survey instrument, the analysis
script, and the thematic coding are available as described in the Data and code availability
statement below.

\section*{CRediT authorship contribution statement}
\textbf{Jesse A. Rodr\'iguez:} Conceptualization, Methodology, Software,
Validation, Formal analysis, Investigation, Resources, Data curation,
Writing --- original draft, Writing --- review \& editing, Visualization,
Supervision, Project administration.

\section*{Declaration of generative AI and AI-assisted technologies in the writing process}
During the preparation of this work the author used Anthropic Claude (Sonnet
and Opus, including the agentic Claude Code system for original code development) for draft editing and code documentation review. After using this
tool, the author reviewed and edited the content as needed and takes full
responsibility for the content of the publication.

\section*{Data and code availability}
The LaTA source code is released under AGPLv3 at
\url{https://github.com/JesseRodriguez/LaTA}. Student-level grading data
are FERPA-protected and cannot be shared; aggregate, de-identified
statistics reported in \S\ref{sec:results} as well as other instruments are available from the author
upon reasonable request.

\section*{Acknowledgments}
The author thanks the School of Mechanical, Industrial, and Manufacturing Engineering at Oregon State University and the Nesbitt Faculty Scholar in Energy Engineering Fund for providing funding that supported this work.

\bibliographystyle{unsrtnat}
\bibliography{references}

\end{document}